\documentclass{article}

\usepackage{graphicx}
\usepackage{caption}
\usepackage{algorithm}
\usepackage{algorithmic}
\usepackage{titlesec}
\usepackage{subcaption}
\usepackage{arxiv}

\usepackage[utf8]{inputenc} 
\usepackage[T1]{fontenc}    
\usepackage{url}            
\usepackage{booktabs}       
\usepackage{amsfonts}       
\usepackage{nicefrac}       
\usepackage{microtype}      
\usepackage{lipsum}		
\usepackage{graphicx}
\usepackage{cite}
\usepackage{doi}
\usepackage[table]{xcolor}  
\usepackage{hyperref}  
\usepackage{tabularx}  
\usepackage{amsmath}
\newcommand\lat{\phi}
\newcommand\lon{\lambda}

\newcommand\distancem{d}

\newcommand\diagm{\distancem_{diag}}
\newcommand\wxm{\distancem_x}
\newcommand\wym{\distancem_y}
\newcommand\agl{AGL}

\newcommand\fov{FOV}
\newcommand\ar{AR}

\newcommand\mercatorx{x}
\newcommand\mercatory{y}
\newcommand\zoom{z}

\newcommand\resolution{res_{x, y}}

\newcommand\scaledres{res'_{x, y}}
\newcommand\mercatorxw{res'_x}
\newcommand\mercatoryw{res'_y}

\newcommand\pixelsize{\Delta p}

\newcommand\movedist{\distancem'}

\newcommand\NVIGdata{NVIG dataset}
\newcommand\Agridata{Agricenter dataset}

\title{TerrAInav Sim: An Open-Source Simulation of UAV Aerial Imaging from Satellite Data}

\author{ \href{https://orcid.org/0009-0007-8288-3386}{\includegraphics[scale=0.06]{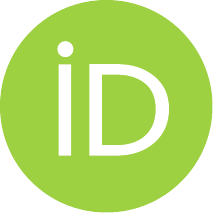}\hspace{1mm}S. Parisa Dajkhosh}\\
	Electrical and Computer Engineering Dept.\\
	University of Memphis\\
	Memphis, TN \\
	\texttt{parisa.d@memphis.edu} \\
	\And
	{\includegraphics[scale=0.06]{orcid.pdf}\hspace{1mm}Peter M. Le} \\
	Electrical and Computer Engineering Dept.\\
	University of Memphis\\
	Memphis, TN \\
	\AND
	{\includegraphics[scale=0.06]{orcid.pdf}\hspace{1mm}Orges Furxhi} \\
	Consultant, Fort Collins CO, USA.\\
	\And
	{\includegraphics[scale=0.06]{orcid.pdf}\hspace{1mm}Eddie L. Jacobs}\\
	Electrical and Computer Engineering Dept.\\
	University of Memphis\\
	Memphis, TN \\
	\texttt{eljacobs@memphis.edu} \\
}

\hypersetup{
	pdftitle={TerrAInav-Sim}
}

\begin{document}
	\maketitle
	
	\begin{abstract}
		Capturing real-world aerial images for vision-based navigation (VBN) is challenging due to limited availability and conditions that make it nearly impossible to access all desired images from any location. The complexity increases when multiple locations are involved. State-of-the-art solutions, such as deploying UAVs (unmanned aerial vehicles) for aerial imaging or relying on existing research databases, come with significant limitations. TerrAInav Sim offers a compelling alternative by simulating a UAV to capture bird's-eye view map-based images at zero yaw with real-world visible-band specifications. This open-source tool allows users to specify the bounding box (top-left and bottom-right) coordinates of any region on a map. Without the need to physically fly a drone, the virtual Python UAV performs a raster search to capture images. Users can define parameters such as the flight altitude, aspect ratio, diagonal field of view of the camera, and the overlap between consecutive images. TerrAInav Sim's capabilities range from capturing a few low-altitude images for basic applications to generating extensive datasets of entire cities for complex tasks like deep learning. This versatility makes TerrAInav a valuable tool for not only VBN but also other applications, including environmental monitoring, construction, and city management. The open-source nature of the tool also allows for the extension of the raster search to other missions. A dataset of Memphis, TN, has been provided along with this simulator. A supplementary dataset is also provided, which includes data from a 3D world generation package for comparison.
	\end{abstract}

	
	\newpage
	\section{Introduction}
	
	\begin{figure}[h]
		\centering
		\includegraphics[width=0.7\linewidth]{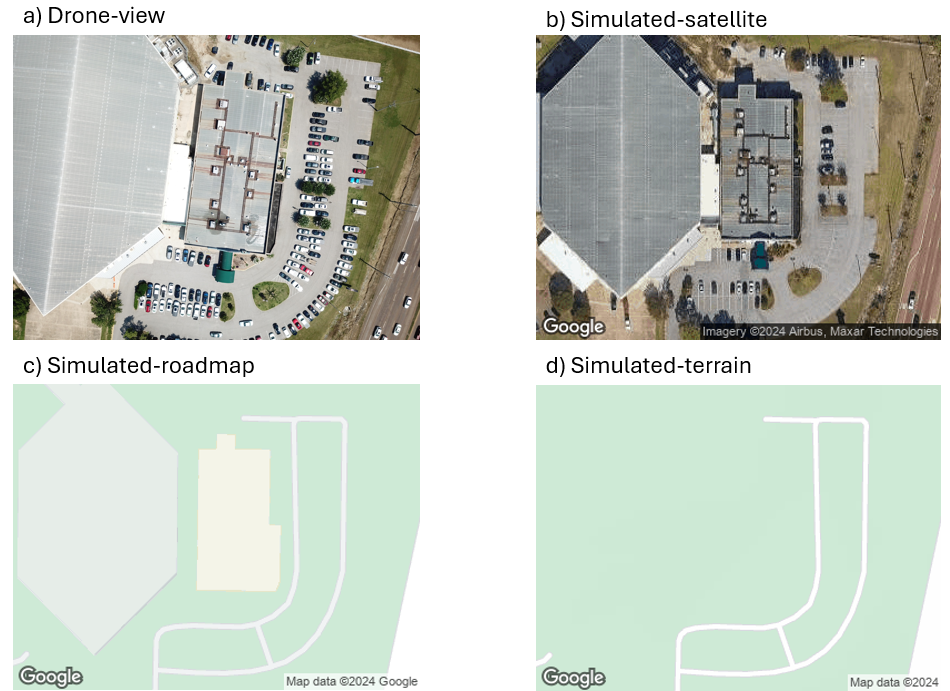}
		\caption{Image captured at coordinates (35.128039, -89.799163) at above ground level altitude ($\agl$) of 126 m. Image captured (a) by a UAV and simulated (b-d) with a field of view ($\fov$) of 78.8 degrees and aspect ratio ($\ar$) of 4:3. Figures b-d represent simulation using three different maptypes.}
		\label{fig:sum}
	\end{figure}
	
	\subsection{Application}
	
	Aerial imagery is crucial for diverse applications, including construction, city
	administration, environmental monitoring, and vision-based navigation. In
	artificial intelligence (AI) and machine learning (ML), aerial images provide
	essential data for training models that power autonomous systems, such as drones
	and self-driving cars, enabling them to navigate by identifying barriers and
	landmarks \cite{aerialnav:Zhang2021}. For environmental monitoring, aerial
	imagery allows for assessing land use changes, tracking deforestation, and
	observing wildlife habitats, all of which are vital for conservation efforts
	\cite{aerialenv:Liang2024}. In construction, these images assist in project
	management by offering real-time site overviews, improving resource allocation,
	and ensuring safety compliance \cite{aerialconst:Liu2024}. City management
	benefits from aerial imagery in urban planning, infrastructure maintenance, and
	disaster response, providing detailed views of city layouts
	\cite{aerialcity:Cao2024}.
	
	Aerial image datasets play a crucial role in machine learning, supporting tasks
	such as object detection, image classification, and segmentation. These datasets enable the identification
	of objects like vehicles, buildings, and vegetation, contributing to urban
	planning, environmental monitoring, and traffic analysis \cite{aerialobjdet}.
	They also support image classification tasks, such as distinguishing between
	land cover types like agricultural, residential, and forested areas, which is
	crucial for resource management
	\cite{long2015fullyconvolutionalnetworkssemantic}. 
	UAV-based remote sensing has been widely applied across domains like agriculture, ecology, and geography. For instance, UAV hyperspectral imaging has been employed for crop yield prediction using machine learning techniques \cite{Zhang2023UAVYield}, highlighting the potential of UAV-based remote sensing in precision agriculture. While hyperspectral imaging provides high spectral resolution, other approaches, such as RGB or multispectral imaging, offer cost-effective alternatives for similar applications.
	In semantic segmentation,
	aerial imagery allows for pixel-level classification, aiding applications such
	as crop monitoring, disaster response, and infrastructure assessment
	\cite{garnot2021panoptic}. Additionally, change detection algorithms use
	temporal aerial imagery to monitor land feature changes over time, essential for
	tracking deforestation, urban expansion, and environmental shifts
	\cite{chaudhuri2012change}. Furthermore, object tracking is another critical
	application of aerial imagery, enabling real-time monitoring of moving objects
	like vehicles and ships in surveillance and security contexts
	\cite{aerialtracking}.

	\subsection{Related Work}
	\subsubsection{Datasets}
	Acquiring real-world aerial images is challenging due to constraints.
	The limited availability of images is a primary issue,
	as it is often impractical to obtain the required aerial views from specific
	locations. This difficulty is compounded when there is a need to cover multiple
	locations on Earth. Deploying Unmanned Aerial Vehicles (UAVs) comes with inherent limitations.
	For instance, flying a UAV to capture aerial images involves logistical complexities, significant time investment, and financial costs. Moreover, UAV operations are subject to regulatory restrictions, such as airspace regulations and flight permissions, as well as environmental factors like weather conditions that can impede flight missions.
	Numerous aerial image datasets are available, serving applications such as
	segmentation and object detection. 
	Real-world aerial images captured by drones are the source of several databases.
	For instance, FloodNet \cite{rahnemoonfar2020floodnet,
		rahnemoonfar2021floodnet}, which comprises 2,343 samples of flooded areas for
	classification, semantic segmentation, and visual question answering (VQA) in
	flood management. The University1652-Baseline dataset \cite{zheng2020university,
		zheng2023uavm} offers satellite-view besides the cross-view (CV) and drone-based
	images.
	
	While these datasets are valuable for many tasks, their usefulness is
	constrained by the parameters set by their authors and may not align with the
	needs of this research. Unlike a real drone mission, these datasets lack the
	flexibility to select preferred locations and altitudes. They also often do not match particular conditions, such as precise locations and camera specifications. Although useful, they may not cover the scenarios needed for
	comprehensive testing or algorithm development. 
	
	A custom imaging simulator
	offers a controlled environment to generate synthetic data, customize
	conditions, and adapt to new models, providing repeatability, safety, and
	cost-effectiveness while addressing issues like incomplete data and challenging
	environments that are difficult to replicate with real-world datasets.
	Moreover, existing UAV-based datasets are often designed for specific applications, such as classification, which limits their adaptability to broader tasks. Even datasets that support multiple applications tend to focus on a narrow set of tasks, restricting their usability in more complex scenarios. Our approach extends these applications by incorporating a wider range of vision-based tasks, including image pattern recognition \cite{mySPIE}, which, when combined with geotags, can contribute to geolocation algorithms. Additionally, segmentation techniques can be applied for road detection and navigation, while image regression, classification, and image-to-image translation open possibilities for applications such as generating synthetic maps for simulation and game design. For example, UAV imagery has been used for road segmentation to enhance autonomous navigation systems \cite{Liu2024roadseg}, and generative models have been applied to aerial imagery to synthesize realistic landscapes for virtual environments \cite{Si2024ganmap}. By expanding the scope of UAV-based image analysis, our dataset aims to bridge these gaps and enable more versatile applications across multiple domains.
	
	\subsubsection{Simulators}
	Flight simulators such as Microsoft Flight Simulator \cite{msflight}, FlightGear Flight Simulator (FGFS) \cite{flightgear}, and Geo-FS \cite{geofs} each come with fundamental limitations that make them unsuitable for imaging research, such as being authentically designed for fixed-wing aircraft flight simulation and limited drone and imaging features that restrict their utility in flight mission imaging research, requiring large-scale raster image collection for machine learning. Microsoft Flight Simulator, while providing a highly detailed and realistic visual experience, is a closed-source platform, preventing any meaningful customization. It is also computationally demanding, requiring powerful hardware, which limits its practicality for high-frequency image capture over large areas.
	FlightGear, despite being open-source, presents similar challenges. It was not designed for imaging applications and lacks native support for drone operations or systematic raster image capture. Significant customization would be required to adapt it, and even then, it suffers from limited terrain resolution and image realism, especially at low altitudes.
	"Geo-FS", while lightweight and accessible as a browser-based simulator, imposes further restrictions: it does not support drone simulations or systematic raster image collection, lacks geotagged data outputs, limits the user’s ability to control camera settings or define extensive missions due to its web-based design and reliance on JavaScript, which hinders compatibility with Python for seamless data integration and manipulation.
	
	Among available flight simulators, FGFS is the only candidate that might be capable of providing comparable data, but not in its default state. It lacks built-in support for precise, automated imaging missions, requiring extensive customization and the development of additional tools to extract usable data. Other simulators, however, are entirely unsuitable, as they are designed purely for piloting experiences and lack any framework for aerial image data generation. Their architectures do not accommodate the level of control, automation, and fidelity necessary for machine learning dataset development, making them completely impractical for this purpose.
	
	To overcome some of these limitations, an alternative solution, such as the MSFS Map Enhancement mod \cite{mapenhancer}, replaces Microsoft Flight Simulator’s default maps with Google Maps imagery, offering more realistic and up-to-date terrain representation. This enhancement along with its being open access, makes it a potential asset for imaging research.
	However, as a third-party modification, Microsoft Flight Simulator is required
	to function and does not operate as a standalone tool, limiting its flexibility
	for large-scale data collection. Despite this, it serves as a valuable reference
	for integrating high-quality map-based imagery for terrAInav Sim.
	
	Developing robust machine learning models for aerial imagery applications
	requires access to large, diverse datasets that ensure both accuracy and
	generalization \cite{sun2017revisiting}. TerrAInav Sim
	\footnote{\url{https://github.com/JacobsSensorLab/TerrAInav-Sim}} addresses these
	needs by providing synthetic aerial datasets that can be used for data
	augmentation, a valuable resource when labeled data is scarce or costly to
	obtain. This lightweight, versatile tool captures geo-tagged, simulated aerial
	images, emulating a UAV capturing bird's-eye view map-based images with
	real-world visible-band specifications.
	By default, the simulated drone is
	positioned at a zero-yaw, facing north, with a 90-degree tilt (looking straight
	down), as shown in Figure~\ref{fig:sum}.
	While this orientation is limited to a top-down perspective, it aligns well with
	many applications that rely on bird’s-eye views for analysis.
	Through TerrAInav Sim, users can simulate diverse
	missions by specifying coordinates for any global location.
	Additionally, for users with the means to collect their
	own drone-view images, TerrAInav extends the capabilities of the University1652
	dataset to any region, enabling tasks that integrate both drone-view and
	satellite-view images for enhanced data collection and analysis.

	\section{Materials and Methods}
	
	In this section, a step-by-step guide will be provided, starting
	with the theoretical calculations required to define various missions. We will
	also delve into the architecture of the program, offering an
	explanation for users to not only utilize, but also advance the program to more sophisticated levels.
	
	\subsection{Mission Definition}
	
	To capture a single image using a drone,
	it is required to collect all necessary materials and resources similar to a real drone flight,
	such as altitude and camera settings. Once a single image is captured, we set a solid foundation
	for more advanced missions. For example, we can program the drone to take
	multiple images at different locations or complete a raster mission by covering
	a specified area, given the bounding box coordinates.
	
	\subsubsection{Single Image Capture}
	
	In order to use Sky-AI Sim, there are some flight and camera parameters that
	have to be known. To capture a single image, these parameters include geodetic latitude $\lat$ and
	longitude $\lon$, above ground level altitude in meters (\(\agl\)), the diagonal field of view
	in degrees (\(\fov\)), and the imaging aspect ratio $\ar$ of the camera. It is
	essential to remember that in this version, the camera tilt is 90 degrees and the drone yaw is 0.
	In satellite simulation, the image pixel size ($\resolution$) and zoom level ($\zoom$)
	respectfully resemble the $\fov$ and $\agl$ in the real world.
	Digital dynamic maps are made possible by loading a small portion of the globe map data at a time,
	allowing for rapid panning and zooming.
	Google Maps uses the Mercator map projection,
	and the world is represented as $256 \times 256$ pixel tiles.
	The zoom levels on the map, ranging from 0 to 23 or higher,
	are represented by a grid of these tiles organized in a pyramidal pattern.
	The scale of the map is determined by each zoom level, which is a numerical scalar.
	For instance, the zoom ranges from 0 to 22 in our real-world simulation if we use the satellite format.
	Tiles show generic information such as continents and oceans at lower zoom levels, resembling flying at a very high altitude.
	On the other hand, more detailed views of streets and buildings are displayed at lower altitudes and bigger zoom levels.
	A grid of \(2^\zoom \times 2^\zoom\) determines the number of tiles at each zoom level ($\zoom$).
	We may use the formula $C_e / 2^\zoom$ level to find the real size of a single tile on a given zoom level,
	where $C_e$ is the circumference of the earth (40,075,017 meters).
	Hence, the number of meters per tile side is obtained \cite{zoom}.

	\begin{table}[H]
		\centering
		\caption{Guide to understand zoom level. The "\# Tiles" column shows the number
			of tiles required to cover the Earth at each zoom level, useful for
			calculating storage needs for pre-generated tiles. "Tile width" column
			indicates the map width in degrees of longitude per square tile at that zoom
			level. The "m / pixels" column lists the meters per pixel at the Equator for
			256-pixel-wide tiles, based on an Earth radius of 6372.7982 km; these values
			should be adjusted by the cosine of the latitude for other regions
			\cite{wiki:zoom, zoom}.}
		\label{tab:zoom-guide}
		\begin{tabular}{lllcl}
			\rowcolor[HTML]{EAECF0}
			\multicolumn{1}{c}{\cellcolor[HTML]{EAECF0}{\color[HTML]{202122}
					\textbf{Level}}} &
			\multicolumn{1}{c}{\cellcolor[HTML]{EAECF0}{\color[HTML]{202122}
					\textbf{\begin{tabular}[c]{@{}c@{}}Tile width\\ (°)\end{tabular}}}} &
			\multicolumn{1}{c}{\cellcolor[HTML]{EAECF0}{\color[HTML]{202122}
					\textbf{\begin{tabular}[c]{@{}c@{}}m / pixel\\ (on Equator)\end{tabular}}}} &
			\multicolumn{1}{l}{\cellcolor[HTML]{EAECF0}\textbf{\begin{tabular}[c]{@{}l@{}}m
						/ tile slide\\ (on Equator)\end{tabular}}} &
			\multicolumn{1}{c}{\cellcolor[HTML]{EAECF0}{\color[HTML]{202122}
					\textbf{\begin{tabular}[c]{@{}c@{}}Examples of\\ areas to
							represent\end{tabular}}}} \\
			\rowcolor[HTML]{F8F9FA}
			{\color[HTML]{202122} 0}
			& {\color[HTML]{202122} 360}
			& {\color[HTML]{202122} 156 543}
			& \multicolumn{1}{l}{\cellcolor[HTML]{F8F9FA}40,075,017}
			& {\color[HTML]{202122} whole world}
			\\
			\rowcolor[HTML]{F8F9FA}
			{\color[HTML]{202122} 1}
			& {\color[HTML]{202122} 180}
			& {\color[HTML]{202122} 78 272}
			& \multicolumn{1}{l}{\cellcolor[HTML]{F8F9FA}20,037,504}
			& {\color[HTML]{202122} }
			\\
			\rowcolor[HTML]{F8F9FA}
			{\color[HTML]{202122} 2}
			& {\color[HTML]{202122} 90}
			& {\color[HTML]{202122} 39 136}
			& \multicolumn{1}{l}{\cellcolor[HTML]{F8F9FA}10,018,752}
			& {\color[HTML]{202122} subcontinental area}
			\\
			\rowcolor[HTML]{F8F9FA}
			{\color[HTML]{202122} 3}
			& {\color[HTML]{202122} 45}
			& {\color[HTML]{202122} 19 568}
			& \multicolumn{1}{l}{\cellcolor[HTML]{F8F9FA}5,009,376}
			& {\color[HTML]{202122} largest country}
			\\
			\rowcolor[HTML]{F8F9FA}
			{\color[HTML]{202122} 4}
			& {\color[HTML]{202122} 22.5}
			& {\color[HTML]{202122} 9 784}
			& \cellcolor[HTML]{F8F9FA}{\color[HTML]{333333} 2,504,688}
			& {\color[HTML]{202122} }
			\\
			\rowcolor[HTML]{F8F9FA}
			{\color[HTML]{202122} 5}
			& {\color[HTML]{202122} 11.25}
			& {\color[HTML]{202122} 4 892}
			& \cellcolor[HTML]{F8F9FA}1,252,344
			& {\color[HTML]{202122} large African country}
			\\
			\rowcolor[HTML]{F8F9FA}
			{\color[HTML]{202122} 6}
			& {\color[HTML]{202122} 5.625}
			& {\color[HTML]{202122} 2 446}
			& \cellcolor[HTML]{F8F9FA}626,172
			& {\color[HTML]{202122} large European country}
			\\
			\rowcolor[HTML]{F8F9FA}
			{\color[HTML]{202122} 7}
			& {\color[HTML]{202122} 2.813}
			& {\color[HTML]{202122} 1 223}
			& \cellcolor[HTML]{F8F9FA}313,086
			& {\color[HTML]{202122} small country, US state}
			\\
			\rowcolor[HTML]{F8F9FA}
			{\color[HTML]{202122} 8}
			& {\color[HTML]{202122} 1.406}
			& {\color[HTML]{202122} 611.496}
			& \cellcolor[HTML]{F8F9FA}156,543
			& {\color[HTML]{202122} US large national park}
			\\
			\rowcolor[HTML]{F8F9FA}
			{\color[HTML]{202122} 9}
			& {\color[HTML]{202122} 0.703}
			& {\color[HTML]{202122} 305.748}
			& \cellcolor[HTML]{F8F9FA}78,272
			& {\color[HTML]{202122} large metropolitan area}
			\\
			\rowcolor[HTML]{F8F9FA}
			{\color[HTML]{202122} 10}
			& {\color[HTML]{202122} 0.352}
			& {\color[HTML]{202122} 152.874}
			& \cellcolor[HTML]{F8F9FA}39,136
			& {\color[HTML]{202122} metropolitan area}
			\\
			\rowcolor[HTML]{F8F9FA}
			{\color[HTML]{202122} 11}
			& {\color[HTML]{202122} 0.176}
			& {\color[HTML]{202122} 76.437}
			& \cellcolor[HTML]{F8F9FA}19,568
			& {\color[HTML]{202122} city}
			\\
			\rowcolor[HTML]{F8F9FA}
			{\color[HTML]{202122} 12}
			& {\color[HTML]{202122} 0.088}
			& {\color[HTML]{202122} 38.219}
			& \cellcolor[HTML]{F8F9FA}9,784
			& {\color[HTML]{202122} town, or city district}
			\\
			\rowcolor[HTML]{F8F9FA}
			{\color[HTML]{202122} 13}
			& {\color[HTML]{202122} 0.044}
			& {\color[HTML]{202122} 19.109}
			& \cellcolor[HTML]{F8F9FA}4,892
			& {\color[HTML]{202122} village, or suburb}
			\\
			\rowcolor[HTML]{F8F9FA}
			{\color[HTML]{202122} 14}
			& {\color[HTML]{202122} 0.022}
			& {\color[HTML]{202122} 9.555}
			& \cellcolor[HTML]{F8F9FA}2,446
			& {\color[HTML]{202122} residential area, airport}
			\\
			\rowcolor[HTML]{F8F9FA}
			{\color[HTML]{202122} 15}
			& {\color[HTML]{202122} 0.011}
			& {\color[HTML]{202122} 4.777}
			& \cellcolor[HTML]{F8F9FA}1,223
			& {\color[HTML]{202122} small road}
			\\
			\rowcolor[HTML]{F8F9FA}
			{\color[HTML]{202122} 16}
			& {\color[HTML]{202122} 0.005}
			& {\color[HTML]{202122} 2.389}
			& \cellcolor[HTML]{F8F9FA}611
			& {\color[HTML]{202122} street}
			\\
			\rowcolor[HTML]{F8F9FA}
			{\color[HTML]{202122} 17}
			& {\color[HTML]{202122} 0.003}
			& {\color[HTML]{202122} 1.194}
			& \cellcolor[HTML]{F8F9FA}306
			& {\color[HTML]{202122} block, park}
			\\
			\rowcolor[HTML]{F8F9FA}
			{\color[HTML]{202122} 18}
			& {\color[HTML]{202122} 0.001}
			& {\color[HTML]{202122} 0.5972}
			& \cellcolor[HTML]{F8F9FA}153
			& {\color[HTML]{202122} some buildings, trees}
			\\
			\rowcolor[HTML]{F8F9FA}
			{\color[HTML]{202122} 19}
			& {\color[HTML]{202122} 0.0005}
			& {\color[HTML]{202122} 0.2986}
			& \cellcolor[HTML]{F8F9FA}76
			& {\color[HTML]{202122} local highway, crossing details}
			\\
			\rowcolor[HTML]{F8F9FA}
			{\color[HTML]{202122} 20}
			& {\color[HTML]{202122} 0.00025}
			& {\color[HTML]{202122} 0.1493}
			& \cellcolor[HTML]{F8F9FA}38
			& {\color[HTML]{202122} mid-sized building}
			\\
			21
			& 0.000125
			& 0.0746
			& \cellcolor[HTML]{F8F9FA}19
			& a house
			\\
			22
			& 0.0000625
			& 0.0373
			& \cellcolor[HTML]{F8F9FA}10
			& a single apartment
		\end{tabular}
	\end{table}
	
	The process of capturing a map, involves understanding the relationship between the
	zoom level, image pixel size, and geolocation bounds. The API allows
	one to specify the zoom level of the map. The zoom level determines the
	scale of the map and, consequently, the resolution of the captured
	image. Higher zoom levels provide more detail, while lower zoom levels
	cover larger areas with less detail. As follows, this section explains how the simulation parameters
	$\agl, \fov, \ar$ are converted to parameters required to capture satellite data
	including $\zoom$ and $\resolution$ \footnote{The functions introduced in "Single Image Capture" section endnotes are located in the "src/utils" folder within the "geo\_helper" package.}.

	\begin{itemize}
		\item \textbf{Map Dimensions in Meters:}
		
		To calculate the image diagonal in meters (\(\diagm\)) given \(\agl\)
		and \(\fov\), use:
		
		\begin{equation}
			\diagm = 2 \cdot \agl \cdot \tan\left(\beta \cdot \fov/2\right)
		\end{equation}
		
		where $\beta$ is a scaler to convert degrees to radians ($\pi / 180$).
		Knowing \(\diagm\), the width of the map along the x and y-axes ($\wxm$, $\wym$) in meters can be calculated as follows
		\footnote{The calculations of land size in meters can be performed using the "get\_map\_dim\_m" function.}:

		\begin{equation}
			\begin{split}
				\wxm =& \diagm \cdot \sin\left(\tan^{-1}(\ar)\right) \\
				\wym =& \diagm \cdot \cos\left(\tan^{-1}(\ar)\right)
			\end{split}
		\end{equation}

		\item \textbf{Determine Geolocation Bounds from Metrics}
		
		Knowing the dimensions in meters, with the benefits of the geopy package
		\cite{geopy}, the bounding box coordinates will be determined
		\footnote{The process of determining geolocation bounds from metric information is performed by the "calc\_bbox\_m" function.}.
		This step can be skipped if the bounding box is already known.
		
		\item \textbf{Determine Zoom Level and Pixel Size from Bounding Box}
		
		The goal of this section is to find the zoom level that maximizes the pixel-wise resolution for the
		specified area. This involves: 1. using the geolocation bounds to
		determine the map's span in degrees; 2. calculating the necessary zoom
		level to fit the entire span within the image dimensions while
		providing the highest resolution. We can achieve the goal of this section,
		by inversing the function explained in the next step.
		
		\item \textbf{Determine Bounding Box from Zoom Level and Pixel Size}
		
		The Google Maps JavaScript API \cite{bbox} getBounds() function provides the latitude and
		longitude boundaries of the visible map area based on the central
		coordinates, zoom level, and map size in pixels ($\resolution$, also
		referred to as the resolution of the map).
		To simulate a geographic area from a central point, zoom level, and map dimensions,
		we calculate the bounding box coordinates using Mercator projection principles.
		The following steps detail this calculation process.
		\footnote{"calc\_bbox\_api" function implements bounding box calculation in Python.}
	\end{itemize}
	\begin{itemize}
		
		\item \textbf{Center Point Offset:} Offsetting the center point from the center of a web Mercator projection tile $(128, 128)$ according to latitude and longitude ($\mercatorx_c$ and $\mercatory_c$) \cite{wiki:mercator}:

		\begin{equation}
			\begin{aligned}
				\mercatorx_c &= 128 + \alpha \cdot \beta \cdot \lon \\
				\mercatory_c &= 128 +  \frac{\alpha}{2} ln\frac{1 + sin(\beta \cdot \lat)}{1 - sin(\beta \cdot \lat)}
			\end{aligned}
			\label{eqn:lat2pnt}
		\end{equation}

		$\alpha$ is a constant to scale the longitude from radians
		to pixels to represents the number of pixels per radians ($256 / 2 \pi$).
		
		\item \textbf{Bounding Box in Mercator Tile:}
		The map's resolution in pixels is scaled to the tile in Mercator
		projection, with the help of a scale factor called pixel size. This
		parameter is calculated from the zoom level ($\zoom$):

		\begin{equation}
			\pixelsize = 2^{-\zoom}
			\label{eqn:pixelsize}
		\end{equation}

		Knowing the scaler ($\pixelsize$), it is also possible to obtain the
		scaled width and height of the region of interest using:

		\begin{equation}
			\scaledres = \resolution \cdot \pixelsize
			\label{eqn:resscale}
		\end{equation}

		where $\scaledres$ is the resolution in pixels mapped from the real world
		to the web mercator projection.
		The bounding box pixels 
		are expnaded by half the width from the central point.
		\item \textbf{Convert to coordinates:}
		Convert pixel values to coordinates using the inverse of Eqn.~\ref{eqn:lat2pnt}
		($pnt$ is the top-left ($TL$) or bottom-right ($BR$) point.):

		\begin{equation}
			\label{pnt2lat}
			\begin{aligned}
				\lon_{pnt} &= (\mercatorx_{pnt} - 128) / (\alpha \cdot \beta)  \\
				\lat_{pnt} &= sin^{-1}(tanh((\mercatory_{pnt} - 128) / (\alpha \cdot \beta)))
			\end{aligned}
		\end{equation}

	\end{itemize}
	
	Knowing how to calculate the bounding box (getting the top left and
	bottom right latitude and longitude from the pixel size and zoom level
	of the geolocation of the center of the image), it is possible to
	reverse the process. In the reverse process, the bounding box is known.
	Hence, the center is known.
	The maximum possible $\resolution$ accessible from the static API is $640 \times 640$.
	With the bounds, finding $\scaledres$ is even easier:
	$\mercatorxw = \mercatorx_{BR} - \mercatorx_{TL}$.
	The same applies to the y-axis. Reversing Eqn. \ref{eqn:pixelsize}:

	\begin{equation}
		\zoom = \lfloor -log_2(max(\mercatorxw, \mercatoryw)/640) \rfloor
	\end{equation}

	Finding the maximum of the resolution's width and height
	ensures the image fits within the selected zoom level.
	Using Eqn. \ref{eqn:pixelsize} and Eqn. \ref{eqn:resscale}, the actual
	resolution of the image can be measured.
	\footnote{ "get\_zoom\_from\_bounds" function finds a zoom level that can maximize the resolution of the captured image by the API}.


	\subsubsection{Raster Image Mission}
	
\begin{figure}[h]
		\centering
		\includegraphics[width=0.7\linewidth]{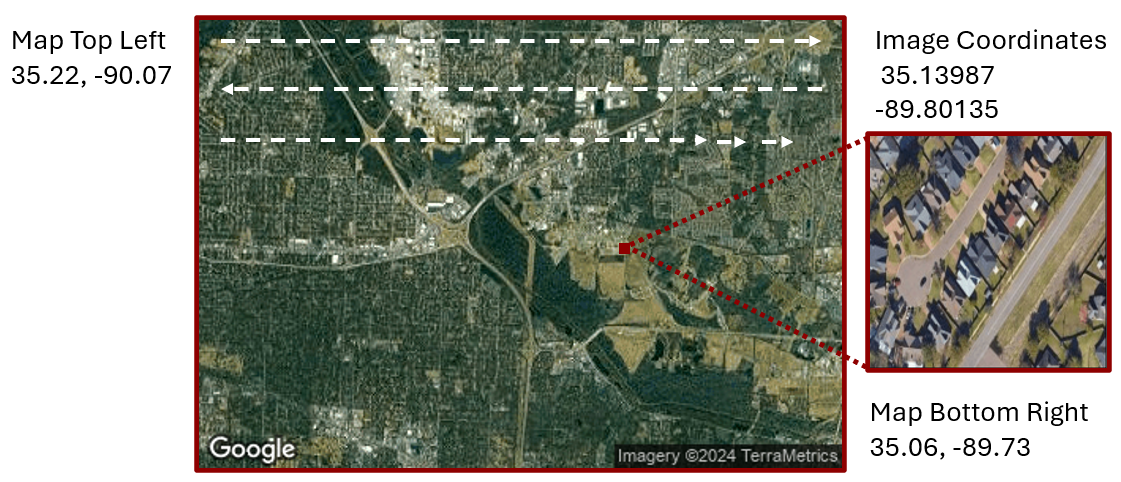}
		\caption{Raster mission visualization within UTM zone 16, provided bounding box coordinates.}
		\label{fig:raster}
	\end{figure}

	Building on single-image capture, more advanced missions can be defined,
	such as raster image collection. The main purpose of TerrAInav Sim is to capture
	raster images of a specified area. In this mission, the objective is to complete
	a zero-tilt raster image capture, as shown in Figure~\ref{fig:raster}.
	The capture begins once the bounding box coordinates of the mission's rectangular region (the map) are set.
	The coordinates are converted to UTM x and y to facilitate metric calculations
	and maintain the mission within a rectangular boundary. To ensure the mission
	remains within one or two UTM zones, it is advisable to restrict its scope to a
	practical distance.
	The first image is centered at the top left coordinates of the map.
	Zoom and image resolution will be calculated from
	the previous section by inputting the camera $\fov$, $\ar$, and the drone's $\agl$.
	By default, images do not overlap, and the distance between consecutive captures
	equals the image width. Overlap can be introduced by adjusting the spacing using
	$\movedist = \distancem (1 - overlap)$ where
	$overlap$ is the proportion of shared width/height to the total and has a value between 0 and 1.
	The mission ends when the bottom right coordinate is captured.
	
	\begin{figure}[htbp]
		\centering
		\includegraphics[width=\linewidth]{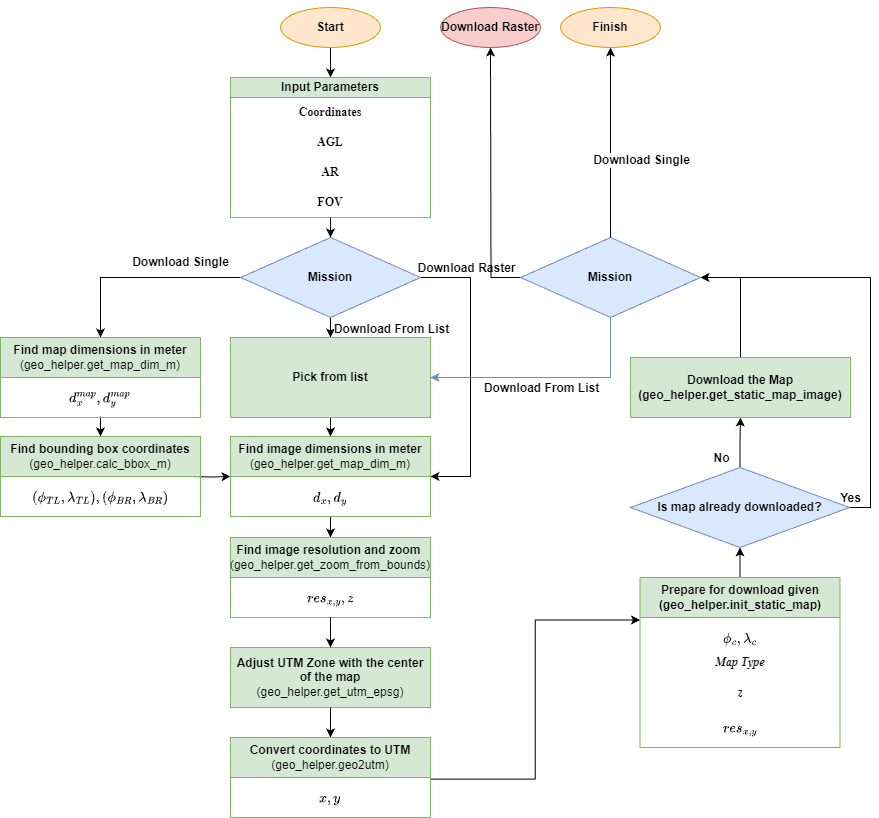}
		\caption{The flowchart to process the map in downloading picture/s, using various missions. Each block includes the name of the function used available in the "src/utils" folder within the "geo\_helper" package in parantheses. The "Download Raster" Section refers to the flowchart in Figure~\ref{fig:rasterflowchart}.}
		\label{fig:mapflowchart}
	\end{figure}
	
	\begin{figure}[htbp]
		\centering
		\includegraphics[width=\linewidth]{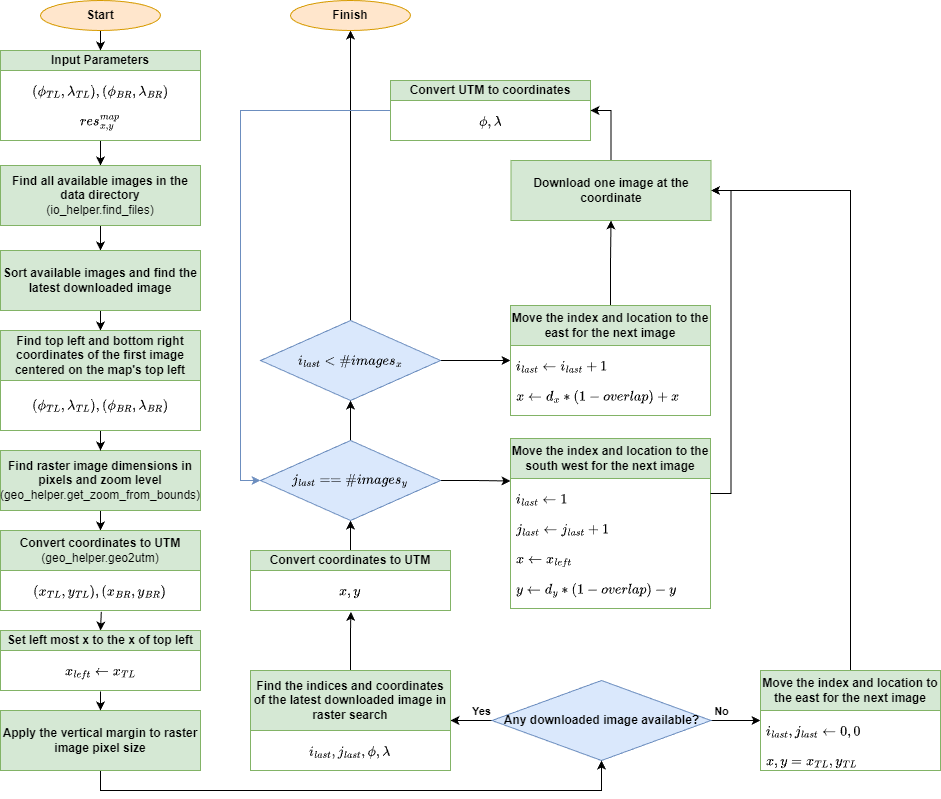}
		\caption{Raster search image capture flowchart.}
		\label{fig:rasterflowchart}
	\end{figure}
	
	\subsection{Machine Learning PreProcessing}
	
\begin{figure}[htbp]
		\centering
		\includegraphics[width=1\linewidth]{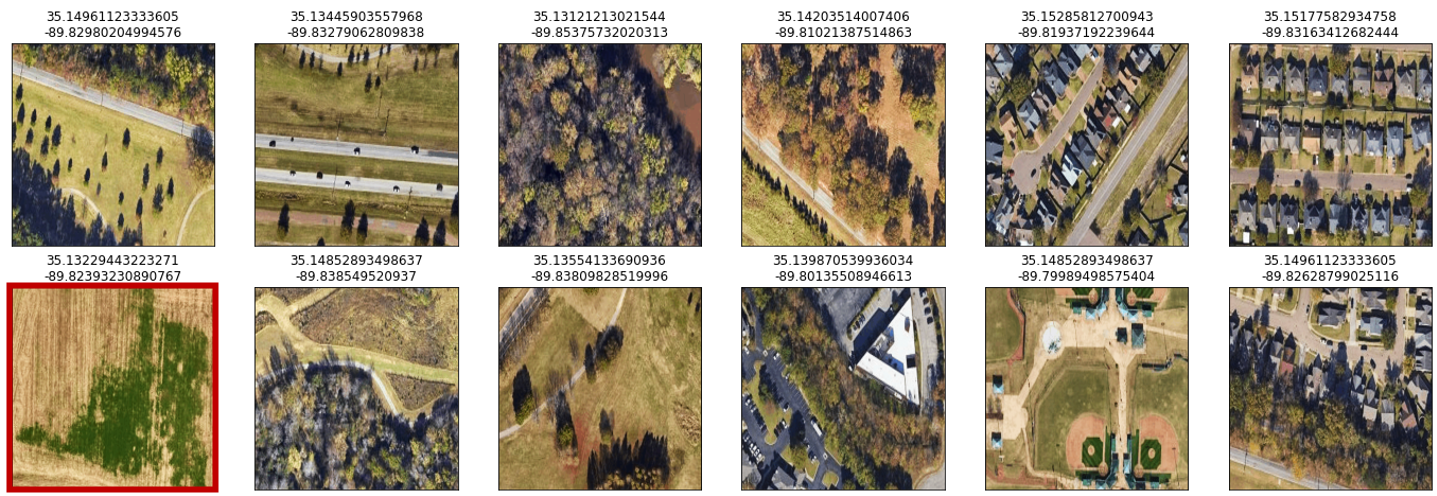}
		\caption{Satellite samples. The image with a red border indicates no significant feature.}
		\label{fig:sample-sat}
	\end{figure}

	Since machine learning requires a substantial amount of data, raster missions
	are valuable for providing this dataset. However, not all raster samples may
	contain useful features for deep learning.
	Additionally, as the altitude decreases, the likelihood of obtaining featureless images increases. Therefore, it becomes increasingly important to filter out images that lack significant features.
	A few examples in Figure~\ref{fig:sample-sat} demonstrate this.
	
\begin{figure}[h]
		\centering
		\includegraphics[width=1\linewidth]{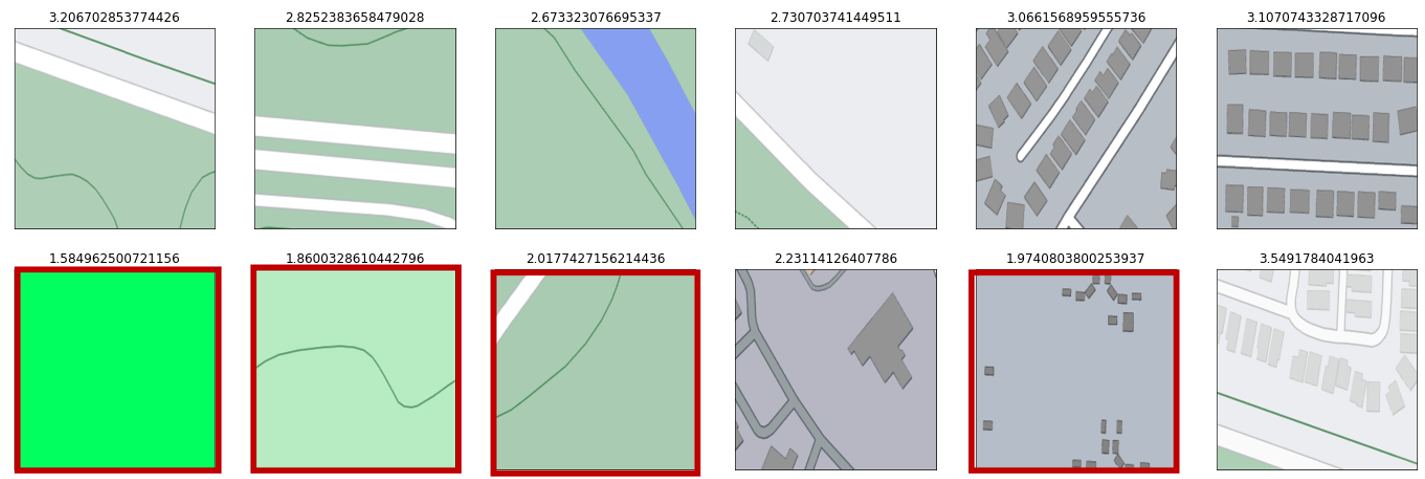}
		\caption{The same images as in Figure~\ref{fig:sample-sat} but in roadmap type. In roadmap, it is easier to detect lack of significant features as marked in red. The entropy of each image is marked on top of it.}
		\label{fig:sample-road}
	\end{figure}

	Shannon entropy can be used to clean data by identifying and removing samples with insignificant features.
	It measures the uncertainty or randomness in data, indicating the complexity of an image.
	High entropy suggests more complexity, while low entropy indicates order \cite{wiki:entropy}.
	To calculate entropy: 1. Convert the image to grayscale; 2. Calculate pixel intensity histogram;
	3. Normalize the histogram to get the probability distribution of pixel intensities.
	4. Calculate the Shannon entropy using:
	
	\begin{equation}
		H = -\sum_{i=0}^{255} p_i \log_2(p_i)
		\label{eqn:entropy}
	\end{equation}
	
	where \( p_i \) is the probability of the pixel intensity \( i \).
	Entropy values help identify samples with fewer significant features.
	Their values are stored in metadata, and the ones below a certain threshold can be removed later.
	By looking at the samples in Figure~\ref{fig:sample-sat}
	using the roadmap mode (as illustrated in Figure~\ref{fig:sample-road})
	it is easier to detect the ones with fewer significant features.
	
	\section{Usage and Results}
	
	\subsection{TerrAInav Dataset}
\begin{figure}
		\centering
		\includegraphics[width=\linewidth]{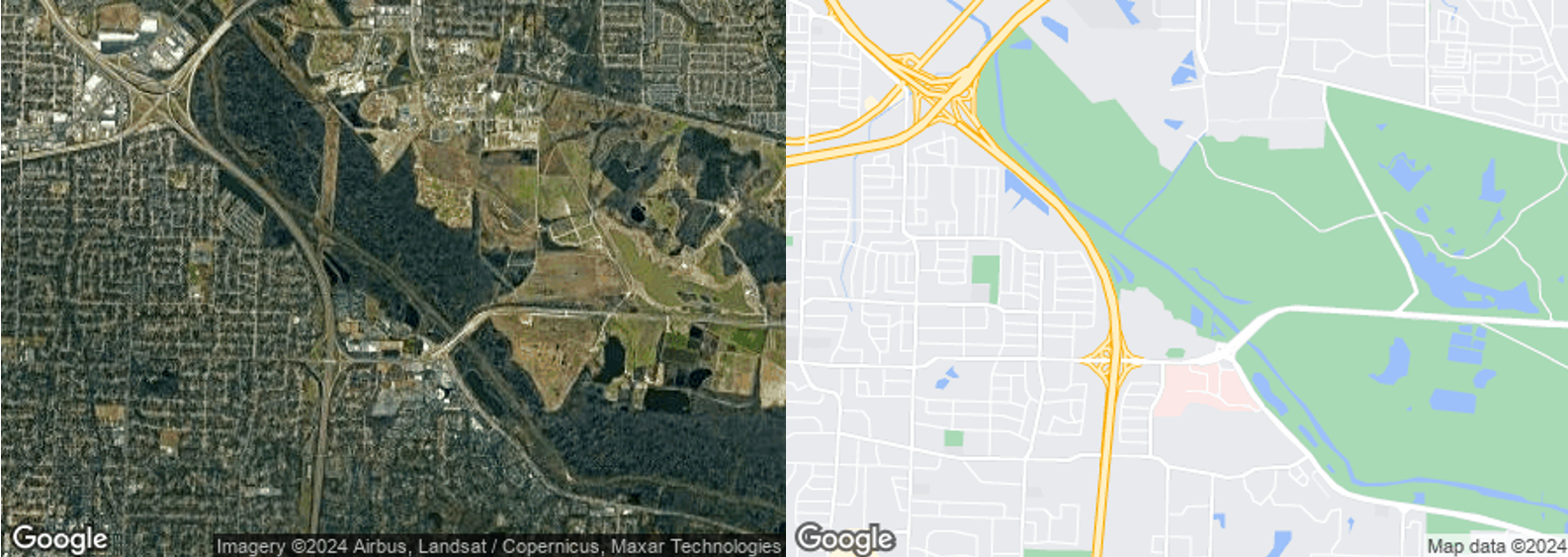}
		\caption{Maps representing the area where the raster data has been collected in "satellite" mode on the left and "roadmap" on the}
		\label{fig:datamap}
	\end{figure}

	Using TerrAInav-Sim, some sample data has been generated and published in this paper
	available in the "dataset" folder. "Memphis\_single\_sample" is an example of
	using the "download\_single" program.
	"random\_samples" folder includes sample data listed in the "sample\_coords.txt".
	This data is collected using "download\_from\_list" program.
	The main dataset includes the raster images of the Memphis Agricenter Area in Figure~\ref{fig:datamap} referred to as \Agridata.
	The images are collected in the rectangular bound between
	top left coordinates of $(35.16, -89.90)$ and bottom right coordinates of
	$(35.115, -89.823)$ using the "download\_raster" program.
	The simulation has been done using a UAV based camera with
	a $\fov$ of 78.8 degrees and $\ar$ of 4:3.
	The flight takes place at a fixed $\agl$ of 120 m.
	There are two sets of 1806 images each in "satellite" and "roadmap" modes without any overlap.
	A bigger dataset
	from Memphis area helped with the project described in \cite{mySPIE}.
	This data is placed in "Memphis" folder under two subfolders "satellite\_0" and "roadmap\_0".
	This dataset is specifically designed to be used for machine learning tasks.
	For ease of access, the images are geotagged; a metadata table is also generated named "meta\_data.csv".
	This file contains 6 columns: 1. "img\_names" has the name of all images available.
	2. "columns" and 3. "rows" consequently refer to the column and row indices of the raster search.
	4. "Lat", 5. "Lon" also refer to the central coordinates of each image.
	6. "Alt" for this dataset stores the zoom level value for each image.
	Some corresponding ML processings will be explained as follows.
	
	\subsection{Supplementary Dataset}
	As previously noted, there are restrictions on publishing data from the API,
	making the TerrAInav datasets themselves, insufficient for machine learning purposes. However, a
	more extensive dataset can be obtained from publicly available road data. In
	light of this, we are offering a larger second dataset that covers the city of
	Memphis. This additional dataset will serve as a valuable resource and can be
	used for comparison with the data generated by TerrAInav Sim in various machine
	learning tasks.
	
	NVIG, or the Night Vision Image Generator, is used to generate terrain for for capturing with sensors created in NVIPM or Night Vision Integrated Performance Model.
	The OpenStreetMap dataset \cite{osm} was used to generate a representation of the Memphis area depicted in Figure \ref{fig:datamap} for NVIG.  The primary data that
	OpenStreetMap provides us are accurate road coordinates. By downloading all the road data within the bounding box coordinates using QGIS program \cite{qgis} tools, we get shape files
	which are then parsed to extract the vectors of road coordinates. The most approachable data to extract after the roads
	are the trees.  By creating a threshold function to mask out the regions of satellite imagery of the bounded Memphis area,
	we can fill the masked regions with random coordinates of varying densities to simulate the presence of trees.
	
	With the road and tree features extracted as series and individual coordinates respectively, the remaining task was generating
	the appropriate road and tree files for NVIG to render the features properly.  After the tools were developed to automate this,
	the addition of roads and trees to the NVIG terrain simulation was quick and efficient.  However, some caveats that resulted
	from this technique were artifacts in the road that manifested as sharp and unnatural spikes in the terrain of which the
	cause has not been discovered. Another problem comes from the way roads are generated in NVIG. The roads are not fitted to
	the terrain; instead, they are formed by points and try to fit to the terrain curvature using a very broad slope parameter.
	A workaround has been to elevate the height of the road so that it is offset some distance above the ground. As most of the imagery
	is aerial, this does not pose a great issue; however, it would be problematic if imagery were to be collected from a
	ground level. Additionally, the road widths are rarely specified in OpenStreetMap. This forces us to use an arbitrary road width
	when it is not provided. As a result, the road width is not a reliable metric for feature extraction with the NVIG terrain.
	
	Generating buildings was also a consideration for the NVIG Agricenter terrain dataset; however, the complexity and effort to implement
	automated building placement in NVIG deterred us from taking that route. The end result is a medium fidelity 3D representation of
	the Agricenter terrain over which custom simulated sensors can collect data. Despite the lack of buildings and the occasional road
	artifacts, the availability of coordinate-accurate roads in a 3D space is powerful in and of itself.
	
\begin{figure}[h]
		\captionsetup[subfigure]{justification=centering}
		\begin{subfigure}[t]{.3\textwidth}
			\centering
			\caption{35.113862, -89.801008}
			\vspace{10pt}
			\includegraphics[width=0.8\linewidth]{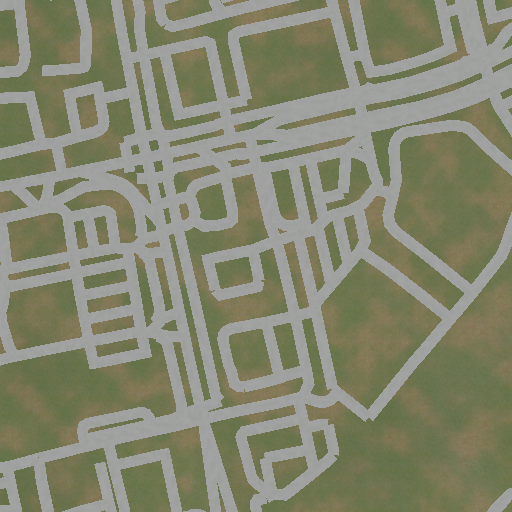}
		\end{subfigure}
		\begin{subfigure}[t]{.3\textwidth}
			\centering
			\caption{35.111531, -89.808043}
			\vspace{10pt}
			\includegraphics[width=0.8\linewidth]{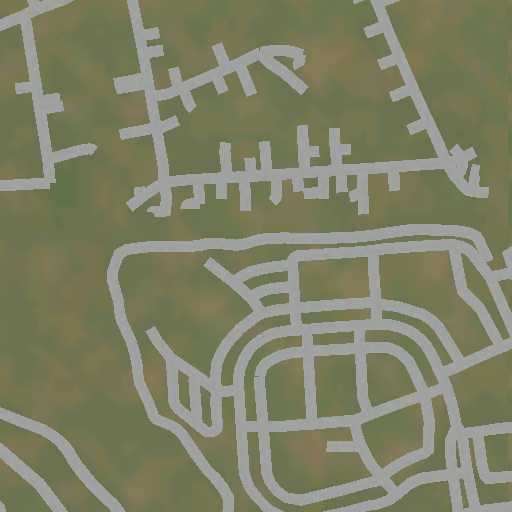}
		\end{subfigure}
		\begin{subfigure}[t]{.3\textwidth}
			\centering
			\caption{35.118524, -89.843216}
			\vspace{10pt}
			\includegraphics[width=0.8\linewidth]{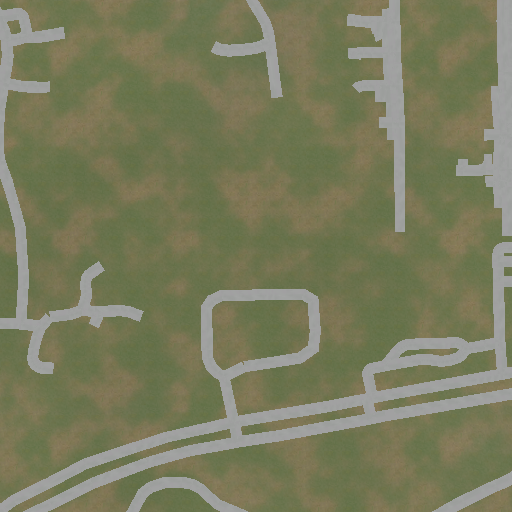}
		\end{subfigure}
		\caption{Sample road data extracted from OpenStreetMap dataset simulated with the NVIPM model. Central coordinates are provided for each sample.}
		\label{fig:nvig}
	\end{figure}

	The bounding box for this data has a top left coordinate of (35.218752, -90.075361), and bottom right (35.064913, -89.730661).
	The images are taken in an $\agl$ of 300 m with horizontal and vertical $\fov$ of 75 and make a total of 3350 images.
	This additional dataset will enhance our ability to generate road data
	and will serve as a key point of comparison with the data from TerrAInav Sim. By
	comparing the two datasets, we can better assess their strengths and
	limitations, ultimately improving the accuracy of our machine
	learning models specifically for road extraction tasks. Figure~\ref{fig:nvig}
	illustrates one sample of this dataset.
	
	\subsection{Usage}
	
	The project consists of two main folders: 1. "src" has all the python codes in it.
	2. "datasets" is the predefined location to store images to, or load the stored data from.
	There are more details about this folder and its content in the dataset section.
	The focus of this section is on the "src" folder and the code structure.
	
	\begin{itemize}
		\item terrainav.data: The data package has three main data modules in it:
		\begin{enumerate}
			\item ImageData: The abstract module to manage image like data.
			\item VBN: The module inherited from ImageData with more specific features for VBN tasks.
			This module can be used to load the \NVIGdata.
			\item TerrAInav: This is the main module to load the TerrAInav data.
			It is inherited from VBN module and has its specific attributes and methods to handle GoogleMap static API.
		\end{enumerate}
		
		\item terrainav.utils: The utility package for managing more generic tasks:
		
		\begin{itemize}
			\item consts: Stores all constant values.
			\item config: Manages input parsed arguments.
			\item geo\_helper: Includes all helper functions for geolocation based calculations.
			Different parts of this module have been explicitly explained in the Missions section.
			\item img\_helper: Consists of all helper functions required to mange images in general.
			\item io\_helper: Helper functions that help with inputting and outputing data is in this package.
			It connects the code to the local directories and the terminal output.
			\item preprocess: Additional image preprocessing tools are provided in this package.
		\end{itemize}
		
		\item terrainav.notebooks: There are two main notebooks available in this directory
		They both take care of processing of the datasets that are currently downloaded.
		"datacleaner" notebook helps with looking at the entropies of random samples
		and preparing the histogram and analysis on the data cleaner threshold selection.
		"preprocess" notebook eases the visualization of applying various preprocessing tools
		on random sample images from the downloaded data.
	\end{itemize}

	\subsubsection{Input Formats for the Argument Parser}
	
	The argument parser for running the script, accepts multiple parameters as input.
	
	\begin{enumerate}
		
		\item \textbf{coords:} The coordinates is entered as a string. It can have different formats:
		\begin{enumerate}
			\item \textbf{Address to a file:} If the input is a string referring to a file,
			the file should contain coordinates and $\agl$ values in the order of
			1. $\lat$, 2. $\lon$, 3. $\agl$ in meters, per line, separated by a space as indicated below.
			When processing a single download file, only the first row will be considered.
			
			\begin{quote}
				$\lat_1$ $\lon_1$ $\agl_1$\\
				$\lat_2$ $\lon_2$ $\agl_2$\\
				\ldots
			\end{quote}
			
			\item \textbf{Formatted String Input:} Alternatively, the input can be provided as a string in one of the following formats:
			
			\begin{itemize}
				\item  \textbf{Raster Coordinates:} Top-left and bottom-right coordinates for raster download. The input format should be:
				
				\begin{quote}
					\texttt{"$\lat_{TL}$\_$\lon_{TL}$\_$\lat_{BR}$\_$\lon_{BR}$\_$\agl$"}
				\end{quote}
				
				For example:
				\begin{quote}
					\texttt{"35.22\_-90.07\_35.06\_-89.73\_400"}
				\end{quote}
				
				In the case of a single download, the top-left coordinate will be used as the central coordinate, and the bottom-right coordinate will be discarded.
				
				\item  \textbf{Central Coordinates and AGL:} The input can also be provided as the central coordinates for a single image capture:
				
				\begin{quote}
					\texttt{"$\lat$\_$\lon$\_$\agl$"}
				\end{quote}
				
			\end{itemize}
		\end{enumerate}
		\item \textbf{fov:} Diagonal field of view in degrees. The default is 78.8.
		\item \textbf{aspect-ratio:} The aspect ratio width and height separated by
		a space, e.g., 4 3 for a 4:3 ratio which is the default value.
		\item \textbf{map\_type:} The type of static map for realistic view is set
		to "satellite" by default. It can be adjusted to "roadmap" or "terrain" as well.
		\item \textbf{dataset:} A string to specify which dataset is going to be used.
		\item The default is "TerrAInav". It can be switched to "VBN" to access and process the Supplementary data.
		\item \textbf{data\_dir:} The directory to read/write the data from/to.
		\item \textbf{img\_size:} After capturing the data, they can get resized to any desired resolution specified by this parameter.
		It takes three parameters:1. width, height, and number of channels (1 for grayscale and 3 for RGB) separated by a space.
		\item \textbf{overlap:} This parameter is set to 0 by default and is only effective for the raster download program.
		It is the proportion of shared width/height to the total and has a value between 0 and 1.
		\item \textbf{batch\_size:} Batch size is only an effective parameter if post processing machine learning dataset is used. It is set to 8 by default.
		\item \textbf{seed:} An initial value used by a random number generator to produce a sequence of numbers, ensuring the same sequence can be reproduced in future runs. The default is 2024 and it can be adjusted if necessary.
	\end{enumerate}
	
	These parameters all have their default values in the "config.py" file located in the "src/utils" directory.
	If other parameters than the default values is desired, specifically for the coordinates, they can be adjusted
	in the command line terminal, as a bash file, or as a "config.json" file.
	The command line has the highest priority. If no values are provided there or in
	a bash script, the "config.json" file will be used. If neither is set, the
	defaults in "config.py" will apply.
	
	\subsection{Post Processing}
\begin{figure}[h]
		\centering
		\includegraphics[width=0.45\linewidth]{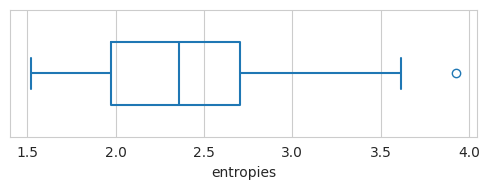}
		\caption{Bar plot showing the entropies of all roadmap images, with most images having an entropy above the 2.1 threshold.}
		\label{fig:entropybar}
	\end{figure}

\begin{figure}[h]
		\centering
		\includegraphics[width=0.45\linewidth]{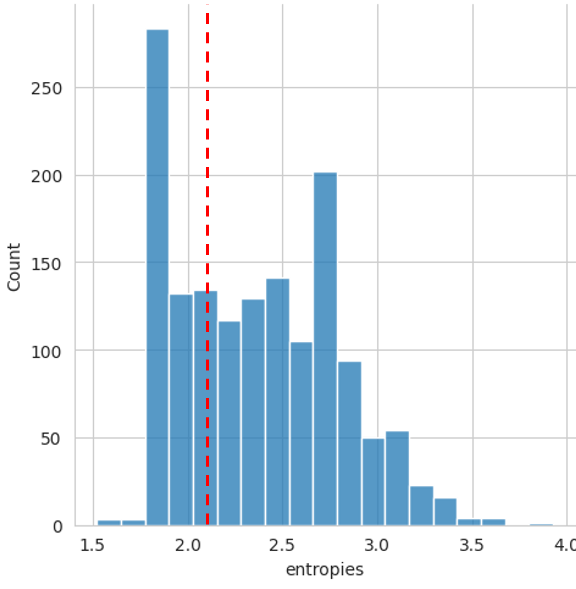}
		\caption{Histogram of image entropies with the red line marking the 2.1 threshold. As shown, a significant portion of data falls below this threshold, but these samples mostly consist of unremarkable features, like green areas with trees and farms. Discarding them is ideal, as the majority of valuable data remains safely above the threshold.}
		\label{fig:entropyhist}
	\end{figure}

	To cleanup the data as an optional step, a threshold entropy is defined.
	The threshold is selected according to the distribution of entropies in all images,
	along with experimental observations provided in Figure~\ref{fig:sample-road}.
	Specifically in the \Agridata, the images below an entropy of 2.1 lack significant details.
	To confirm this value's reliability and ensure it doesn't filter out substantial
	valuable data, a barplot and histogram of entropies for all available roadmap
	images are shown in Figures~\ref{fig:entropybar} and
	\ref{fig:entropyhist}.
	This ensures most data is above the threshold, making it safe to discard images below $2.1$.
	TerrAInav performs data cleaning based on the map type. If the map type
	is "satellite" or anything other than "roadmap", it first checks for "roadmap"
	data in the data directory. If available, images are filtered according to
	"roadmap" images with an entropy above the default threshold (2.1). If no
	roadmap data exists, images are filtered based on their original map type.
	
	After the data is cleaned using the "cleanup\_data" method, the "config\_dnn"
	method can be called to prepare the dataset for a machine learning task. In this
	process, the labels are the central latitude and longitude of each image, which
	are then converted to UTM and normalized using a standard scaler. The data is
	then split into three sets: train, validation, and test. At this stage,
	the cleaned data is transferred to a Keras dataset \cite{kerasdataset}, where it is preprocessed and
	divided into batches. Data augmentation is then applied, including random
	rotation, flip, and zoom by default. Additional options like random contrast and
	brightness adjustments are also available if needed.

\section{Discussion}

TerrAInav Sim opens the door to numerous applications, but it does have limitations that could affect its utility in certain scenarios. One notable limitation arises from the reliance on the Google Maps Static API, which prevents the determination of the exact date of the imagery. This can lead to discrepancies in temporal accuracy, especially in applications where precise timestamping is crucial, such as change detection in environmental monitoring or tracking land use over time in agricultural applications.
This limitation, while not critical for most of TerrAInav Sim's potential
applications, is still an important area for future improvement. To address this limitation, a potential solution is to integrate additional data sources that provide time-stamped satellite imagery. For instance, services like Google Earth Engine
\cite{api:ee} offer access to historical satellite imagery with precise timestamps, which could be leveraged through their APIs. However, their archive still has their own limitations.
Integrating these services into TerrAInav Sim would significantly improve the temporal accuracy of the imagery, making the tool more suitable for applications that require historical comparisons, event tracking, or temporal analysis.

Another limitation lies in the idealized nature of the simulation, which does not account for the complexities and inaccuracies typical of real-world drone missions. Factors such as sensor faults, varying environmental conditions, and obstacles like buildings or trees can significantly impact the quality of images captured during actual operations. For example, in tasks like urban mapping or vegetation analysis, these inaccuracies could introduce errors in feature extraction or object recognition. Currently, TerrAInav Sim provides a controlled, distortion-free environment, but this does not reflect the unpredictability of real-world operations. To mitigate this, incorporating error models simulating sensor noise, dynamic obstacles, and environmental factors (e.g., wind or lighting changes) could better represent real-world conditions. This would allow users to assess how their systems would perform in diverse, unpredictable scenarios, which is particularly important for applications like environmental monitoring or disaster response.

The current simulator also uses a high-altitude image capture and cropping method to simulate low-altitude drone behavior. While the proposed approach provides a reasonable approximation, it does not replicate the distortions and misalignments experienced during actual low-altitude flights. This discrepancy can be particularly problematic in applications requiring precise 3D modeling or when operating in densely built-up areas where buildings can create severe perspective distortions. To address this, future versions of the simulator could implement custom distortion models that account for terrain elevation, building height, and the specific angle of capture. This would allow for more realistic simulations of low-altitude operations, improving the accuracy of applications like urban modeling or emergency response planning.

Furthermore, there is a trade-off between image quality and processing speed. High-quality images require more computational resources, which can increase processing time, potentially limiting the simulator's utility in real-time applications, such as autonomous navigation or obstacle avoidance in dynamic environments. On the other hand, faster image acquisition and processing can lead to a reduction in image fidelity, which may compromise the accuracy of object detection and map generation. In future versions, a dynamic adjustment system that optimizes the balance between image quality and processing speed, perhaps based on the specific needs of the task at hand (e.g., prioritizing speed for real-time navigation vs. quality for detailed mapping), could provide a more flexible and practical solution for users.

Lastly, the current image capture setup, which limits yaw to zero and tilt to 90 degrees, restricts the available angles for data capture. This limitation is particularly restrictive in applications that require a full 360-degree view for accurate environmental modeling or surveillance. Expanding the simulator to include dynamic adjustments for both yaw and tilt would allow for a broader range of angles, enabling more comprehensive data capture for tasks like infrastructure inspection, where capturing multiple angles of a structure is essential for thorough analysis.

\section{Conclusion}
TerrAInav Sim's functionality allows users to customize various parameters,
including flight altitude, camera aspect ratio, diagonal field of view, and
image overlap. This flexibility supports a wide range of applications, from
capturing low-altitude images for simple tasks to generating comprehensive
datasets for complex projects. The ability to quickly and efficiently capture large volumes of images while utilizing minimal memory makes it an ideal candidate for deep learning. Its potential applications span various fields, including classification, pattern recognition, localization, and image-to-image translation, showcasing just a few of its capabilities in deep learning.
By offering a faster and simpler alternative to commercial flight simulators, TerrAInav Sim accelerates experimentation and validation processes, making it an invaluable resource for future advancements in UAV-based geospatial analysis, environmental monitoring, and autonomous navigation research.

In addition to its versatility, TerrAInav Sim's open-source nature encourages
further development and adaptation to other missions beyond its initial design.
The tool's ability to produce extensive datasets without the need for physical
UAV deployment represents a significant advancement in dataset generation for VBN algorithm developments.
TerrAInav Sim presents a compelling solution to the challenges of aerial image
capture, offering a powerful and flexible tool for a variety of applications,
and opening new avenues for research and development in VBN
and beyond.

By providing a customizable, cost-effective tool for generating diverse image
datasets, TerrAInav Sim lowers barriers to entry for research in autonomous
navigation, environmental monitoring, and urban planning. Its open-source nature
encourages continued development, making it a valuable asset for advancing
machine learning and remote sensing applications.

	\bibliographystyle{plain}
	\bibliography{skyai}  

\begin{thebibliography}{10}

\bibitem{bbox}
{C}alculate bounding box of static google maps image --- stackoverflow.com.
\newblock
  \url{https://stackoverflow.com/questions/44784839/calculate-bounding-box-of-static-google-maps-image}.
\newblock [Accessed 26-08-2024].

\bibitem{osm}
{O}pen{S}treet{M}ap --- openstreetmap.org.
\newblock \url{https://www.openstreetmap.org/#map=5/38.01/-95.84}.
\newblock [Accessed 26-08-2024].

\bibitem{api:ee}
{P}ython {I}nstallation  |  {G}oogle {E}arth {E}ngine  |  {G}oogle for
  {D}evelopers --- developers.google.com.
\newblock
  \url{https://developers.google.com/earth-engine/guides/python\_install}.
\newblock [Accessed 26-08-2024].

\bibitem{zoom}
{U}nderstanding {Z}oom {L}evel in {M}aps and {I}magery ---
  support.plexearth.com.
\newblock
  \url{https://support.plexearth.com/hc/en-us/articles/6325794324497-Understanding-Zoom-Level-in-Maps-and-Imagery}.
\newblock [Accessed 26-08-2024].

\bibitem{wiki:zoom}
{Z}oom levels - {O}pen{S}treet{M}ap {W}iki --- wiki.openstreetmap.org.
\newblock \url{https://wiki.openstreetmap.org/wiki/Zoom\_levels}.
\newblock [Accessed 26-08-2024].

\bibitem{chaudhuri2012change}
Yifang Ban and Osama Yousif.
\newblock Change detection techniques: A review.
\newblock pages 19--43, 11 2016.

\bibitem{aerialcity:Cao2024}
Yejun Cao, Xiwen Yu, and Fengling Jiang.
\newblock Application of 3d image technology in rural planning.
\newblock {\em ACM Transactions on Asian and Low-Resource Language Information
  Processing}, 23(6):1–13, June 2024.

\bibitem{geopy}
Geopy Contributors.
\newblock Geopy: Geocoding library for python, 2024.
\newblock Version 2.3.0.

\bibitem{msflight}
Microsoft Corporation.
\newblock Microsoft flight simulator.
\newblock [Accessed 04-11-2024].

\bibitem{mySPIE}
S.~Parisa Dajkhosh, Orges Furxhi, C.~Kyle Renshaw, and Eddie~L. Jacobs.
\newblock {Aerial image feature mapping using deep neural networks}.
\newblock In Kannappan Palaniappan and Gunasekaran Seetharaman, editors, {\em
  Geospatial Informatics XIV}, volume PC13037, page PC1303701. International
  Society for Optics and Photonics, SPIE, 2024.

\bibitem{mapenhancer}
Derekhe.
\newblock {G}it{H}ub - derekhe/msfs2020-map-enhancement --- github.com.
\newblock \url{https://github.com/derekhe/msfs2020-map-enhancement}, 2024.
\newblock [Accessed 16-01-2025].

\bibitem{flightgear}
FlightGear Developers.
\newblock Flightgear flight simulator.
\newblock [Accessed 04-11-2024].

\bibitem{Zhang2023UAVYield}
Yahui Guo, Yi~Xiao, Fanghua Hao, Xuan Zhang, Jiahao Chen, Kirsten {de Beurs},
  Yuhong He, and Yongshuo~H. Fu.
\newblock Comparison of different machine learning algorithms for predicting
  maize grain yield using uav-based hyperspectral images.
\newblock {\em International Journal of Applied Earth Observation and
  Geoinformation}, 124:103528, 2023.

\bibitem{aerialobjdet}
Jan Gąsienica-Józkowy, Mateusz Knapik, and Boguslaw Cyganek.
\newblock An ensemble deep learning method with optimized weights for
  drone-based water rescue and surveillance.
\newblock {\em Integrated Computer-Aided Engineering}, pages 1--15, 01 2021.

\bibitem{aerialenv:Liang2024}
Chen-Wei Liang, Zhong-Chun Zheng, and Ting-Nong Chen.
\newblock Monitoring landfill volatile organic compounds emissions by an
  uncrewed aerial vehicle platform with infrared and visible-light cameras,
  remote monitoring, and sampling systems.
\newblock {\em Journal of Environmental Management}, 365:121575, August 2024.

\bibitem{Liu2024roadseg}
Ruyi Liu, Junhong Wu, Wenyi Lu, Qiguang Miao, Huan Zhang, Xiangzeng Liu,
  Zixiang Lu, and Long Li.
\newblock A review of deep learning-based methods for road extraction from
  high-resolution remote sensing images.
\newblock {\em Remote Sensing}, 16(12), 2024.

\bibitem{aerialconst:Liu2024}
Wenjin Liu, Lijuan Zhou, Shudong Zhang, Ning Luo, and Min Xu.
\newblock A new high-precision and lightweight detection model for illegal
  construction objects based on deep learning.
\newblock {\em Tsinghua Science \& Technology}, 29(4):1002–1022, August 2024.

\bibitem{long2015fullyconvolutionalnetworkssemantic}
Jonathan Long, Evan Shelhamer, and Trevor Darrell.
\newblock Fully convolutional networks for semantic segmentation, 2015.

\bibitem{wiki:mercator}
Peter Osborne.
\newblock {The Mercator Projections}, November 2013.
\newblock {Supplements: Maxima files and Latex code and figures}.

\bibitem{qgis}
{QGIS.org}.
\newblock Qgis geographic information system.
\newblock \url{http://qgis.org}, YEAR.
\newblock Open Source Geospatial Foundation Project.

\bibitem{rahnemoonfar2020floodnet}
Maryam Rahnemoonfar, Tashnim Chowdhury, Argho Sarkar, Debvrat Varshney, Masoud
  Yari, and Robin Murphy.
\newblock Floodnet: A high resolution aerial imagery dataset for post flood
  scene understanding.
\newblock {\em arXiv preprint arXiv:2012.02951}, 2020.

\bibitem{rahnemoonfar2021floodnet}
Maryam Rahnemoonfar, Tashnim Chowdhury, Argho Sarkar, Debvrat Varshney, Masoud
  Yari, and Robin~Roberson Murphy.
\newblock Floodnet: A high resolution aerial imagery dataset for post flood
  scene understanding.
\newblock {\em IEEE Access}, 9:89644--89654, 2021.

\bibitem{garnot2021panoptic}
Vivien Sainte Fare~Garnot and Loic Landrieu.
\newblock Panoptic segmentation of satellite image time series with
  convolutional temporal attention networks.
\newblock {\em ICCV}, 2021.

\bibitem{wiki:entropy}
Claude~E. Shannon.
\newblock A mathematical theory of communication.
\newblock {\em Bell System Technical Journal}, 27(3):379--423, 1948.

\bibitem{Si2024ganmap}
Jongwook Si and Sungyoung Kim.
\newblock Gan-based map generation technique of aerial image using residual
  blocks and canny edge detector.
\newblock {\em Applied Sciences}, 14(23), 2024.

\bibitem{sun2017revisiting}
Chen Sun, Abhinav Shrivastava, Saurabh Singh, and Rahul Gupta.
\newblock Revisiting unreasonable effectiveness of data in deep learning era.
\newblock {\em Proceedings of the IEEE International Conference on Computer
  Vision}, pages 843--852, 2017.

\bibitem{geofs}
Xavier Tassin.
\newblock {G}eo{F}{S} - {F}ree {O}nline {F}light {S}imulator --- geo-fs.com.
\newblock [Accessed 04-11-2024].

\bibitem{kerasdataset}
Keras Team.
\newblock {K}eras documentation: {D}atasets --- keras.io.
\newblock \url{https://keras.io/api/datasets/}.
\newblock [Accessed 03-09-2024].

\bibitem{aerialtracking}
Shao-Yu Yang, Hsu-Yung Cheng, and Chih-Chang Yu.
\newblock Real-time object detection and tracking for unmanned aerial vehicles
  based on convolutional neural networks.
\newblock {\em Electronics}, 12(24), 2023.

\bibitem{aerialnav:Zhang2021}
Xupei Zhang, Zhanzhuang He, Zhong Ma, Zhongxi Wang, and Li~Wang.
\newblock Llfe: a novel learning local features extraction for uav navigation
  based on infrared aerial image and satellite reference image matching.
\newblock {\em Remote Sensing}, 13(22):4618, November 2021.

\bibitem{zheng2023uavm}
Zhedong Zheng, Yujiao Shi, Tingyu Wang, Jun Liu, Jianwu Fang, Yunchao Wei, and
  Tat-seng Chua.
\newblock Uavm'23: 2023 workshop on uavs in multimedia: Capturing the world
  from a new perspective.
\newblock In {\em Proceedings of the 31st ACM International Conference on
  Multimedia}, pages 9715--9717, 2023.

\bibitem{zheng2020university}
Zhedong Zheng, Yunchao Wei, and Yi~Yang.
\newblock University-1652: A multi-view multi-source benchmark for drone-based
  geo-localization.
\newblock {\em ACM Multimedia}, 2020.

\end{thebibliography}
	
\end{document}